\documentclass[conference]{IEEEtran}

\usepackage{cite}
\usepackage{amsmath,amssymb,amsfonts}
\usepackage{algorithmic}
\usepackage{graphicx}
\usepackage{textcomp}
\usepackage{xcolor}
\usepackage{url}
\usepackage{hyperref}
\usepackage{tikz}
\usetikzlibrary{positioning, arrows.meta, shapes.geometric, fit, backgrounds}
\usepackage{booktabs}
\usepackage{enumitem}
\usepackage{tcolorbox}
\usepackage{microtype}

\hypersetup{
    colorlinks=true,
    linkcolor=blue!60!black,
    citecolor=blue!60!black,
    urlcolor=blue!60!black
}

\definecolor{downstreamblue}{RGB}{59,130,246}
\definecolor{upstreamamber}{RGB}{245,158,11}
\definecolor{volatilered}{RGB}{239,68,68}
\definecolor{surfacegray}{RGB}{248,248,252}

\tcbset{
  defbox/.style={
    colback=surfacegray,
    colframe=gray!40,
    fonttitle=\bfseries\small,
    boxrule=0.4pt,
    arc=3pt,
    left=4pt, right=4pt, top=3pt, bottom=3pt,
    fontlower=\small
  }
}

\begin{document}

\title{Conversation Tree Architecture: A Structured Framework
for Context-Aware Multi-Branch LLM Conversations}

\author{
  \IEEEauthorblockN{Pranav Hemanth}
  \IEEEauthorblockA{
    Pandemonium Research\\
    pranavh2004@gmail.com
  }
  \and
  \IEEEauthorblockN{Sampriti Saha}
  \IEEEauthorblockA{
    Pandemonium Research\\
    sampritisaha@gmail.com
  }
}

\maketitle

\begin{abstract}
Large language models (LLMs) are increasingly deployed for extended,
multi-topic conversations, yet the flat, append-only structure of current
conversation interfaces introduces a fundamental limitation: all context
accumulates in a single unbounded window, causing topically distinct threads
to bleed into one another and progressively degrade response quality. We
term this failure mode \textit{logical context poisoning}. In this paper,
we introduce the \textbf{Conversation Tree Architecture} (CTA), a
hierarchical framework that organizes LLM conversations as trees of
discrete, context-isolated nodes. Each node maintains its own local context
window; structured mechanisms govern how context flows between parent and
child nodes -- \textit{downstream} on branch creation and \textit{upstream}
on branch deletion. We additionally introduce \textbf{volatile nodes},
transient branches whose local context must be selectively merged upward or
permanently discarded before purging. We formalize the architecture's primitives, characterize
the open design problems in context flow, relate our framework to prior
work in LLM memory management, and describe a working prototype
implementation. The CTA provides a principled foundation for structured
conversational context management and extends naturally to multi-agent
settings.
\end{abstract}

\begin{IEEEkeywords}
large language models, conversational AI, context management,
hierarchical memory, branching conversations, multi-agent systems
\end{IEEEkeywords}

\textit{Note: This paper presents a conceptual framework, architectural
formalization, and prototype implementation. Systematic empirical
evaluation is ongoing and will be reported in subsequent work.}

\section{Introduction}

Large language models have become primary tools for knowledge work,
creative collaboration, software engineering, and research assistance.
As their capabilities have grown, so has the complexity of the tasks
users bring to them. A typical modern LLM session may span dozens of
conversational turns across multiple distinct topics: a user might begin
with a high-level design question, pursue a debugging tangent, revisit
the design, and then explore an entirely separate domain all within a
single conversation.

The current standard interface for LLM conversations is a flat,
chronologically ordered list of message pairs sharing a single context
window that handles this complexity poorly. Earlier turns are never
discarded; they remain in the context and influence subsequent responses
regardless of relevance. As conversations grow, models increasingly
receive context that is internally inconsistent in terms of topic,
abstraction level, and intent. The result is a systematic degradation
in response quality that is difficult for users to diagnose or correct.

We characterize this failure mode as \textbf{logical context poisoning}:
the progressive corruption of conversational coherence caused not by
model error, but by the structural mismanagement of context. The
phenomenon has empirical support: Liu et al.~\cite{liu2024lost}
demonstrate that LLM performance degrades significantly when relevant
information is displaced toward the middle of a long context window;
Hsieh et al.~\cite{hsieh2025context} show that degradation occurs even
when retrieval is perfect, implicating context length itself as a source
of noise; and recent applied research has characterized the effect under
the term ``context rot''~\cite{chroma2025rot}.

Existing user-facing mitigations are inadequate. Starting a new
conversation discards all prior context. Manual summarization is lossy
and labor-intensive. System prompt re-anchoring is fragile and addresses
symptoms rather than structure. There is currently no mechanism within
mainstream LLM interfaces for users to manage context in a hierarchical,
structured manner that mirrors the natural branching topology of complex
thought.

This paper makes the following contributions:

\begin{enumerate}[leftmargin=1.2em, itemsep=1pt]
  \item We introduce and formally define \textbf{logical context
    poisoning} as the primary failure mode of linear LLM conversation
    interfaces.
  \item We propose the \textbf{Conversation Tree Architecture} (CTA),
    a formal framework organizing conversations as directed trees of
    context-isolated nodes.
  \item We define and characterize the key primitives: conversational
    units, branch nodes, local context windows, downstream context
    passing, upstream context merging, cross-node context passing,
    and volatile nodes.
  \item We identify and articulate the open design problems in each
    context flow operation as concrete research questions.
  \item We describe a working prototype implementation at \\
    \url{https://the-conversation-tree.vercel.app/app} and situate the CTA
    within related work in LLM memory management and multi-agent systems.
\end{enumerate}

\section{Background and Related Work}

The CTA sits at the intersection of several active research areas. We
review the most directly relevant prior work and articulate how the CTA
differs from and extends each.

\subsection{Context Window Length and Degradation}

The transformer architecture~\cite{vaswani2017attention} processes input
as a flat sequence with attention over all prior tokens. As context
grows, the signal-to-noise ratio of the context window decreases.
Liu et al.~\cite{liu2024lost} demonstrate a ``lost in the middle''
effect: performance degrades significantly when relevant information is
not positioned at the beginning or end of the context.
Hsieh et al.~\cite{hsieh2025context} show that this degradation holds
even under perfect retrieval conditions, suggesting that the volume of
irrelevant context is itself harmful. Applied research at
Chroma~\cite{chroma2025rot} characterizes the user-facing phenomenon
as ``context rot''. The CTA addresses this at the structural level by
preventing irrelevant context from accumulating in the first place,
rather than attempting to retrieve signal from within a noisy window.

\subsection{Hierarchical and Tiered LLM Memory}

The closest architectural antecedent to the CTA is
MemGPT~\cite{packer2023memgpt}, which proposes virtual context
management inspired by operating system memory hierarchies. MemGPT
maintains a main context (analogous to RAM) and external storage
(analogous to disk), moving information between tiers based on
relevance. The CTA differs in two key respects: first, MemGPT manages
memory within a single linear conversation, whereas the CTA imposes a
structural reorganization of the conversation itself; second, MemGPT's
flow decisions are made autonomously, whereas the CTA treats branching
as a deliberate, user-initiated act reflecting the user's own
understanding of topical boundaries.

Most directly related to the CTA is ContextBranch~\cite{chickmagalur2025contextbranch},
a concurrent work that applies version control semantics to LLM conversations,
formalizing four primitives -- checkpoint, branch, switch, and inject.
A controlled experiment across 30 software engineering scenarios demonstrates
that branching reduces context size by 58.1\% and improves response quality
with large effect sizes on focus ($d=0.80$) and context awareness ($d=0.87$).
The CTA and ContextBranch share the core observation that conversation
isolation prevents context poisoning. However, they differ in three important
respects. First, ContextBranch targets exploratory software engineering
workflows specifically, whereas the CTA is a general-purpose conversational
framework applicable to research, creative work, and multi-agent settings.
Second, the CTA introduces \textbf{volatile nodes}, a formally defined
ephemeral branch type with a mandatory merge-or-purge lifecycle, which has
no analog in ContextBranch. Third, the CTA raises the open problem of
\textit{insertion positioning} during upstream merging (chronological versus
end-append), a design dimension ContextBranch does not address. The two
frameworks are thus complementary: ContextBranch provides empirical
validation of the branching hypothesis that the CTA formalizes at a broader
architectural level.

Rezazadeh et al.~\cite{rezazadeh2024tree} propose dynamic tree memory
representations for LLMs, organizing stored memories as hierarchical
schemas. Their work focuses on the representation of \textit{stored
knowledge} across sessions, whereas the CTA focuses on the structure
of \textit{active conversation} within and across sessions.
Hu et al.~\cite{hu2024hiagent} introduce hierarchical working memory
for long-horizon agent tasks, decomposing tasks into subgoal-structured
chunks -- structurally analogous to our branch hierarchy, though designed
for autonomous agents rather than human-in-the-loop conversations.

A broader research program known as Hierarchical Context Management
(HCM)~\cite{liu2025hcm} organizes context into nested levels with distinct
retention policies, employing techniques such as context folding, level-based
promotion and demotion, and chunked summarization across temporal and spatial
scales. HCM systems create \textit{vertical} memory hierarchies in which
content ascends or descends between tiers based on relevance scores or
temporal decay. The CTA is structurally orthogonal: it creates a
\textit{lateral} branching hierarchy organized around how conversation is
structured, not how memory is tiered. Critically, HCM flow decisions are
driven autonomously by the system, whereas the CTA treats branching as a
deliberate, user-initiated act. The upstream merge and volatile node
primitives of the CTA have no direct analog in HCM frameworks.

\subsection{Conversational Memory Across Sessions}

Several recent systems address persistent memory across LLM sessions.
MemoryBank~\cite{zhong2024memorybank} stores timestamped dialogue
entries and retrieves them via semantic search.
Maharana et al.~\cite{maharana2024longterm} evaluate LLM agents on
long-term conversational memory tasks spanning multiple sessions.
A-Mem~\cite{xu2025amem} proposes a dynamic memory system linking units
through semantic relationships. Supermemory~\cite{daga2025supermemory}
achieves state-of-the-art results on LongMemEval~\cite{wu2024longmemeval}
through relational versioning and temporal grounding.

These systems address the orthogonal problem of what to
\textit{remember} across sessions. The CTA addresses how to
\textit{structure} the conversation during a session, such that the
right information is naturally scoped to the right context. The two
approaches are complementary.

\subsection{Tree-Structured Reasoning and Multi-Agent Systems}

Tree of Thoughts~\cite{yao2023tot} introduces tree-structured search
over LLM-generated reasoning steps, enabling deliberate exploration
of multiple reasoning paths. While structurally related, ToT operates
within a single inference step and concerns model-side reasoning rather
than user-conversation structure. AutoGen~\cite{wu2023autogen} enables
multi-agent LLM applications through structured inter-agent conversation
patterns. The CTA's node-and-flow abstraction naturally extends to such
multi-agent pipelines, as we discuss in Section~\ref{sec:future}.

\section{Problem Formalization}

\subsection{The Linear Conversation Model}

We formalize the current standard conversation model as follows. A
\textbf{linear conversation} $\mathcal{C}$ is a sequence of
conversational units:
\[
  \mathcal{C} = \langle u_1, u_2, \ldots, u_n \rangle
\]
where each \textbf{conversational unit} $u_i = (h_i, r_i)$ is an
ordered pair consisting of a human message $h_i$ and a model response
$r_i$. At turn $n$, the model receives as input the concatenation of
all prior units:
\[
  \mathrm{input}_n = \bigoplus_{i=1}^{n-1}(h_i \oplus r_i) \oplus h_n
\]
where $\oplus$ denotes sequence concatenation. The context window
$W_n = \mathrm{input}_n$ grows monotonically and is never pruned or
restructured.

\subsection{Logical Context Poisoning}

\begin{tcolorbox}[defbox, title=Definition 1: Logical Context Poisoning]
\textit{Logical context poisoning} is the progressive degradation of
model response quality caused by the accumulation of topically
inconsistent, abstraction-mismatched, or task-irrelevant content
within a single shared context window. It is a structural failure mode
distinct from model error: outputs become misaligned not because the
model is wrong, but because the context is an incoherent mixture of
distinct conversational threads.
\end{tcolorbox}

Logical context poisoning is characterized by three concurrent conditions:
\begin{itemize}[leftmargin=1.2em, itemsep=1pt]
  \item The context window contains content from multiple distinct topical
    threads $T_1, T_2, \ldots, T_k$ with $k > 1$;
  \item The content of thread $T_i$ is not uniformly relevant to the
    current thread $T_j$ ($i \neq j$); and
  \item The model has no mechanism to identify or discount content from
    $T_i$ when generating a response in the context of $T_j$.
\end{itemize}

\subsection{Limitations of Existing Mitigations}

Users currently employ three primary strategies, all inadequate:

\textbf{New conversation:} Discards all prior context, including
genuinely relevant content, and forces manual re-establishment of
grounding.

\textbf{Manual summarization:} Labor-intensive, lossy, and does not
scale to complex multi-topic sessions.

\textbf{System prompt re-anchoring:} Fragile -- earlier content remains
in the context window and is not removed, only overridden and is
inaccessible to non-technical users.

\section{The Conversation Tree Architecture}

\subsection{Formal Definition}

\begin{tcolorbox}[defbox, title=Definition 2: Conversation Tree]
A \textit{conversation tree} $\mathcal{T}$ is a directed, rooted tree
$\mathcal{T} = (V, E, r, W)$ where:
\begin{itemize}[leftmargin=1em, itemsep=1pt, topsep=2pt]
  \item $V$ is a finite set of \textbf{conversation nodes}
  \item $E \subseteq V \times V$ is a set of directed parent-child edges
  \item $r \in V$ is the \textbf{root node}
  \item $W : V \rightarrow \mathcal{W}$ maps each node to its
    \textbf{local context window} $w_v \in \mathcal{W}$
\end{itemize}
Each node $v \in V$ maintains an independent context window $w_v$.
Information passes between nodes only through defined flow operations.
\end{tcolorbox}

\subsection{Conversation Nodes and Local Context Windows}

Each node $v \in V$ corresponds to a focused conversational task or
topic, defined at branching time by the user. A node accumulates
conversational units over its lifecycle:
\[
  w_v = \langle u_1^v, u_2^v, \ldots, u_{n_v}^v \rangle
\]
The model operating within $v$ receives only $w_v$ as context at
inference time, plus any content passed from the parent via the
downstream flow operation. Local context windows are mobile -- they
can be partially or fully transferred to other nodes through the flow
operations defined below.

\subsection{Downstream Context Passing}

When a user creates a child node $v_c$ from parent $v_p$, a downstream
context passing operation initializes $w_{v_c}$:
\[
  w_{v_c}^{(0)} = \phi_{\downarrow}(w_{v_p},\, \theta)
\]
where $\phi_{\downarrow}$ is a \textbf{downstream selection function}
parameterized by $\theta$. This presents three open design problems:

\begin{enumerate}[leftmargin=1.2em, itemsep=1pt]
  \item \textbf{Relevance selection}: which units $u_i^{v_p}$ are
    relevant to the child's intended purpose?
  \item \textbf{Compression level}: should units be passed verbatim,
    as extractive summaries, or as abstractive compressions?
  \item \textbf{Poisoning avoidance}: excessive downstream transfer
    reintroduces context poisoning within the child.
\end{enumerate}

\subsection{Upstream Context Merging}

When child $v_c$ is deleted, an upstream merging operation incorporates
relevant content from $w_{v_c}$ into $w_{v_p}$:
\[
  w_{v_p}^{(\mathrm{new})} = \psi_{\uparrow}(w_{v_p},\, w_{v_c},\, \delta)
\]
where $\psi_{\uparrow}$ is an \textbf{upstream merge function}
parameterized by $\delta$ specifying merge position. This presents
three open design problems:

\begin{enumerate}[leftmargin=1.2em, itemsep=1pt]
  \item \textbf{Relevance filtering}: which content from $w_{v_c}$
    merits preservation -- by novelty, task-relevance, or
    user-specified importance?
  \item \textbf{Condensation granularity}: should merged content be
    verbatim units, condensed summaries, or structured extracts?
  \item \textbf{Insertion positioning}: should merged content be
    appended to the end of $w_{v_p}$, or inserted at the chronological
    position of the branch point?
  \item \textbf{Chunked and staggered insertion}: rather than inserting
    all merged content at once, insights from $w_{v_c}$ may be introduced
    incrementally -- interleaved with the parent's continuing conversation
    stream. This raises questions about insertion ordering, pacing, and
    whether the model can coherently integrate context that arrives in
    discrete, non-contiguous fragments.
\end{enumerate}

\subsection{Cross-Node Context Passing}

Cross-node context passing is the general class of operations by which
context migrates between any two nodes $(v_i, v_j) \in V \times V$:
\[
  w_{v_j}^{(\mathrm{new})} = \xi(w_{v_i},\, w_{v_j},\, \rho)
\]
where $\xi$ is a general context transfer function and $\rho$ specifies
direction, selection criteria, and merge policy. This subsumes
downstream and upstream flows and additionally encompasses lateral
transfer between sibling or non-adjacent nodes.

\subsection{Volatile Nodes}

\begin{tcolorbox}[defbox, title=Definition 3: Volatile Node]
A \textit{volatile node} $v \in V$ is a branch node annotated with a
\texttt{volatile} flag. It exists only for the duration of a session
and carries no persistent state. At session termination or explicit
deletion, $w_v$ must either be merged upstream via $\psi_{\uparrow}$
or purged entirely. If purged, $w_v$ is irretrievably discarded.
\end{tcolorbox}

Volatile nodes enable exploratory conversational threads (hypotheses,
tangents, debugging sessions, alternative framings) where the user
investigates without committing to preservation. The merge-or-purge
decision creates a principled checkpoint: the user explicitly decides
whether the branch's content has value worth returning to the parent.
The lifecycle of a volatile node is:
\[
  \texttt{create} \;\rightarrow\; \texttt{interact} \;\rightarrow\;
  \texttt{delete} \;\rightarrow\; \{\texttt{merge} \mid \texttt{purge}\}
\]

\subsection{Architectural Diagram}

\begin{figure*}[t]
\centering
\begin{tikzpicture}[
  node distance=1.4cm and 2.6cm,
  every node/.style={font=\small},
  cnode/.style={circle, draw=gray!60, fill=white, thick,
    minimum size=0.75cm},
  vnode/.style={circle, draw=volatilered!80, fill=volatilered!10,
    thick, minimum size=0.75cm},
  cw/.style={rounded corners=3pt, draw=gray!40, fill=surfacegray,
    minimum width=1.2cm, minimum height=0.45cm, font=\scriptsize},
  down/.style={-Stealth, thick, color=downstreamblue},
  up/.style={-Stealth, thick, color=upstreamamber, dashed},
]
\node[cnode] (root) {$r$};
\node[cnode, below left=of root]  (A)  {$A$};
\node[cnode, below right=of root] (B)  {$B$};
\node[cnode, below left=of A]     (A1) {$A_1$};
\node[vnode, below right=of A]    (A2) {$A_2$};
\node[vnode, below=of B]          (B1) {$B_1$};

\node[cw, right=0.15cm of root] {$w_r$};
\node[cw, left=0.15cm  of A]    {$w_A$};
\node[cw, right=0.15cm of B]    {$w_B$};
\node[cw, left=0.15cm  of A1]   {$w_{A_1}$};
\node[cw, right=0.15cm of A2]   {$w_{A_2}$};
\node[cw, right=0.15cm of B1]   {$w_{B_1}$};

\draw[down] (root) -- (A)
  node[midway, left, color=downstreamblue, font=\scriptsize]
  {$\phi_{\downarrow}$};
\draw[down] (root) -- (B);
\draw[down] (A) -- (A1);
\draw[down] (A) -- (A2);
\draw[down] (B) -- (B1);

\draw[up] (A2) to[bend right=28] (A)
  node[midway, right, color=upstreamamber, font=\scriptsize]
  {$\psi_{\uparrow}$};
\draw[up] (B1) to[bend left=22] (B);

\begin{scope}[shift={(6,-1.8)}]
  \draw[gray!30, rounded corners=4pt]
    (-0.3,0.5) rectangle (3.2,-2.0);
  \node[font=\small\bfseries] at (1.45,0.2) {Legend};
  \node[cnode, minimum size=0.4cm] (ln) at (0,-0.2) {};
  \node[right=0.12cm of ln, font=\scriptsize] {Normal node};
  \node[vnode, minimum size=0.4cm] (lv) at (0,-0.7) {};
  \node[right=0.12cm of lv, font=\scriptsize] {Volatile node};
  \draw[down,-Stealth] (-0.2,-1.2) -- (0.45,-1.2)
    node[right, font=\scriptsize] {Downstream ($\phi_{\downarrow}$)};
  \draw[up,-Stealth]   (-0.2,-1.55) -- (0.45,-1.55)
    node[right, font=\scriptsize] {Upstream ($\psi_{\uparrow}$)};
\end{scope}

\end{tikzpicture}
\caption{A representative Conversation Tree. Nodes $A_2$ and $B_1$
  are volatile (red). Blue solid edges: downstream passing
  $\phi_{\downarrow}$. Amber dashed edges: upstream merging
  $\psi_{\uparrow}$. Each node holds an independent local context
  window $w_v$.}
\label{fig:architecture}
\end{figure*}
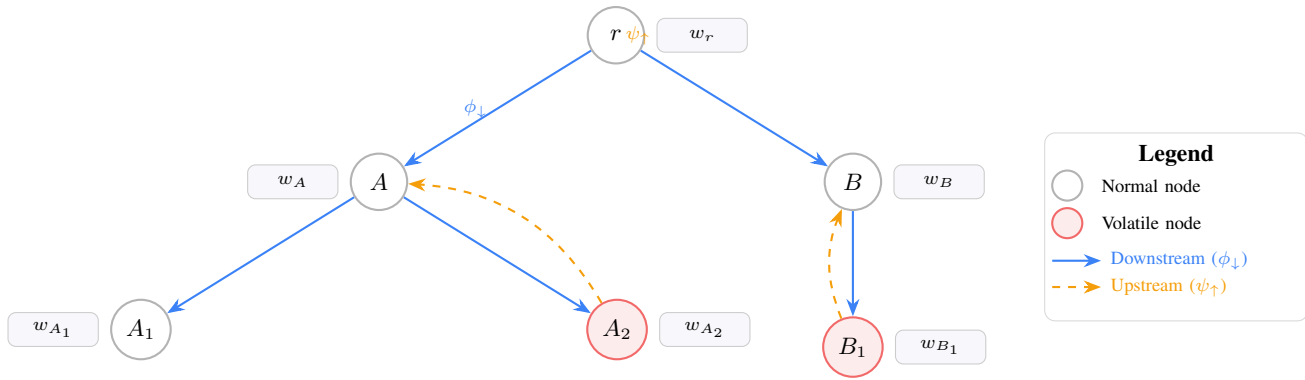

Fig.~\ref{fig:architecture} illustrates the CTA with a representative
tree. Blue solid edges indicate downstream context passing; amber dashed
edges indicate upstream merging; red nodes are volatile.

\section{Prototype Implementation}
\label{sec:prototype}

We have developed a working prototype of the CTA, publicly accessible
at \url{https://the-conversation-tree.vercel.app/app}. The prototype
demonstrates the core structural primitives and serves as a
proof-of-concept for the feasibility of the approach.

\subsection{Implementation Overview}

The prototype is implemented as a web application using React and the
React Flow library for interactive tree visualization, with LLM inference
provided via the Groq and Gemini APIs. Key implemented features include:

\begin{itemize}[leftmargin=1.2em, itemsep=1pt]
  \item \textbf{Node creation and branching}: users create child nodes
    from any existing node, establishing the tree structure through an
    interactive drag-and-drop graph interface
  \item \textbf{Full context isolation}: each node maintains an independent
    conversation history; the model receives only the current node's local
    context window at inference time, with no leakage between sibling nodes
  \item \textbf{Downstream context passing}: on branch creation, the user
    selects whether to pass the full parent context or start the child with
    a clean window; the selected content is injected as the child's initial
    context
  \item \textbf{Volatile node designation}: nodes can be flagged as volatile
    at creation time, rendering them with distinct visual styling and
    triggering a merge-or-purge prompt on deletion
  \item \textbf{Interactive visualization}: the conversation tree is rendered
    as a live, color-coded graph with blue edges for downstream flow and
    amber edges for upstream merges
\end{itemize}

\subsection{Current Limitations}

The current prototype implements the structural primitives of the CTA but
defers intelligent flow algorithms to future work. Downstream passing
supports full-context or no-context transfer; selective relevance filtering
and compression are not yet implemented. Upstream merging is currently
manual, the user specifies what to carry back, and automatic condensation
or chronological insertion positioning is not implemented. Lateral cross-node
passing between non-adjacent nodes is not yet supported. These limitations
constitute the primary research and engineering agenda described in
Section~\ref{sec:goals}.

\section{Significance and Applications}
\label{sec:significance}

\subsection{User-Facing Applications}

Three representative use cases illustrate the breadth of applicability:

\textbf{Research and analysis.} A researcher maintains a primary
analytical thread as the root node. Volatile child nodes are created
to explore individual papers or follow citation chains; useful findings
are merged upstream and dead ends are purged, leaving the root coherent
throughout. Beyond exploration, the researcher can also fork the main
thread into homogeneous sub-threads, each pursuing a complementary
angle of the same research question, allowing parallel investigation
that ultimately converges through upstream merges rather than forcing
all lines of inquiry into a single increasingly cluttered context.

\textbf{Software engineering.} A software engineer maintains a
high-level architecture design as the root node. Debugging sessions
branch from it. The root is not contaminated by low-level details;
when debugging concludes, a structured summary is merged upstream.

\textbf{Creative writing.} A writer maintains a narrative planning
thread as the root node. Volatile branches allow experimentation with
alternative scenes or voices without committing to their preservation.

\subsection{Multi-Agent Generalization}

The node-and-flow abstraction extends naturally to multi-agent LLM
systems~\cite{wu2023autogen}. In a multi-agent configuration, each
node $v \in V$ may be associated with a distinct agent. Cross-node
context passing then corresponds to inter-agent communication, and
the tree structure provides a principled topology for organizing agent
collaboration with focused local context and selective information
propagation. This direction is elaborated as a research goal in
Section~\ref{sec:future}.



\section{Research Goals and Open Problems}
\label{sec:goals}

\subsection{Formal Computational Model}

The immediate objective is to formalize the CTA as a precise
computational model, specifying data structures and algorithms for each
primitive operation, defining invariants that must hold throughout the
tree's lifecycle, and characterizing the complexity of context flow
operations as a function of tree depth and node context size.

\subsection{Intelligent Downstream Context Passing}

Key research questions for $\phi_{\downarrow}$: Can semantic similarity
between the child's initial prompt and parent context units serve as a
reliable relevance signal? What is the optimal compression ratio for
downstream passing? Can $\phi_{\downarrow}$ be learned end-to-end from
human feedback on branch quality?

\subsection{Upstream Merge Strategies}

The insertion positioning question (end-append versus chronological
insertion at the branch point) is particularly consequential and has
not, to our knowledge, been studied in any prior work on LLM memory
management. We conjecture that chronological insertion will outperform
end-append on tasks requiring causal coherence, but this remains to be
validated empirically. A further open question is whether \textit{chunked
and staggered insertion} (interleaving merged content incrementally into
the parent's ongoing conversation stream rather than injecting it
atomically) better preserves conversational coherence than bulk
insertion at a single point.

\subsection{Empirical Evaluation}

Relevant evaluation dimensions include:

\begin{itemize}[leftmargin=1.2em, itemsep=1pt]
  \item \textbf{Response quality}: does the CTA reduce measurable
    quality degradation on multi-topic conversational tasks?
  \item \textbf{Task completion}: does tree-structured conversation
    improve completion rates on complex, multi-step tasks?
  \item \textbf{User-reported coherence}: do users report higher
    perceived coherence in tree-structured conversations?
  \item \textbf{Context efficiency}: does the CTA reduce total tokens
    consumed for equivalent task outcomes?
\end{itemize}

\section{Future Scope}
\label{sec:future}

\subsection{Multi-Agent Architectures}

The CTA's node-and-flow abstraction extends naturally to multi-agent
LLM systems~\cite{wu2023autogen}. In a multi-agent configuration, each
node $v \in V$ may be associated with a distinct specialized agent. The
downstream flow $\phi_{\downarrow}$ then corresponds to task delegation
with selective context seeding; the upstream merge $\psi_{\uparrow}$
corresponds to result aggregation with relevance filtering; and cross-node
passing $\xi$ corresponds to peer-to-peer inter-agent communication. The
tree topology provides a principled structure for organizing agent
collaboration hierarchies that flat, unstructured multi-agent
frameworks do not provide natively. Formalizing CTA semantics for
multi-agent orchestration, and evaluating whether tree-structured agent
communication improves task completion on long-horizon benchmarks, is a
primary direction for future work.

\subsection{Towards a Structured Second Brain}

At a longer time horizon, the CTA represents a foundational layer for
AI-augmented knowledge work. The analogy to Molecular
Notes~\cite{martin2022molecular} is instructive: just as Molecular
Notes organizes static knowledge into Sources, Atoms, and Molecules
with explicit linking structure, the CTA organizes \textit{generative}
knowledge work (active conversation) into a structured hierarchy with
explicit flow operations. The upstream merge is the CTA's equivalent of
atom extraction: distilling essential insight from exploratory branches
and integrating it into a more permanent knowledge structure. A longer-term
vision is a system in which the accumulated topology of a user's
conversation tree, across sessions and projects, begins to function as a
genuine second brain: generative, organized, and coherent.

\section{Conclusion}

We have presented the Conversation Tree Architecture, a hierarchical
framework for organizing LLM conversations as trees of context-isolated
nodes with structured flow operations between them. The CTA addresses
logical context poisoning -- the systematic degradation of conversational
coherence caused by the linear, append-only structure of current LLM
interfaces through principled mechanisms for context isolation,
directed flow, and selective merging.

The architecture introduces a set of novel primitives (volatile nodes,
upstream merging with chronological and chunked insertion positioning,
intelligent downstream selection) that distinguish it from prior work
in LLM memory management and the concurrent ContextBranch system.
A working prototype demonstrates the feasibility of the approach.
The open problems identified in context flow algorithm design constitute
a concrete research agenda for subsequent work, with multi-agent
generalization and structured second-brain systems representing the
longer-term vision outlined in Section~\ref{sec:future}.




\end{document}